\pdfoutput=1
\documentclass[10pt,twocolumn,letterpaper]{article}

\usepackage{cvpr}              

\usepackage[dvipsnames]{xcolor}

\definecolor{cvprblue}{rgb}{0.21,0.49,0.74}
\usepackage[pagebackref,breaklinks,colorlinks,citecolor=cvprblue]{hyperref}

\usepackage{multirow}
\usepackage{color}
\usepackage[accsupp]{axessibility}  
\definecolor{gray}{RGB}{60,60,60}

\def\paperID{31} 
\def\confName{VAND2.0}
\def\confYear{2024}

\title{LogicAL: Towards logical anomaly synthesis for\\ unsupervised anomaly localization}

\author{Ying Zhao\\
Ricoh Software Research Center (Beijing) Co., Ltd.\\
{\tt\small zy$\_$deepwhite$\_$zy@hotmail.com}
}

\begin{document}
\maketitle
\begin{abstract}
Anomaly localization is a practical technology for improving industrial production line efficiency. Due to anomalies are manifold and hard to be collected, existing unsupervised researches are usually equipped with anomaly synthesis methods. However, most of them are biased towards structural defects synthesis while ignoring the underlying logical constraints. To fill the gap and boost anomaly localization performance, we propose an edge manipulation based anomaly synthesis framework, named LogicAL, that produces photo-realistic both logical and structural anomalies. We introduce a logical anomaly generation strategy that is adept at breaking logical constraints and a structural anomaly generation strategy that 
complements to the structural defects synthesis. We further improve the anomaly localization performance by introducing edge reconstruction into the network structure. Extensive experiments on the challenge MVTecLOCO, MVTecAD, VisA and MADsim datasets verify the advantage of proposed LogicAL on both logical and structural anomaly localization.

\end{abstract}    
\section{Introduction}
\label{sec:intro}

\begin{figure*}
  \centering
  \setlength{\abovecaptionskip}{0.cm}
    \includegraphics[width=\linewidth]{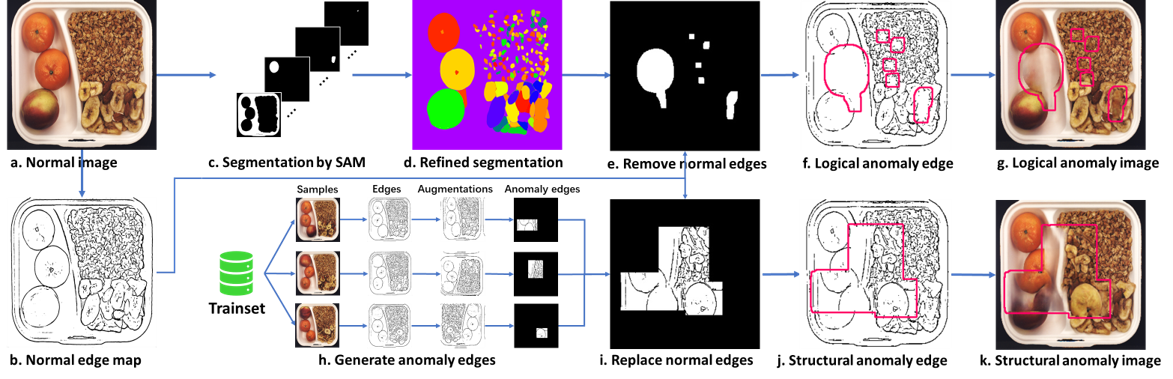}
  \caption{\textbf{Framework of LogicAL.} LogicAL converts a normal image into an anomaly image based on edge manipulation. (c) to (g) show that it generates logical anomaly image by removing normal edges from the selected semantic regions. It syntheses structural anomaly image by replacing normal edges in arbitrary regions with the augmented edges that are sampled from trainset, as shown in (h) to (k).}
  \label{fig:2}
  \vspace{-5mm}
\end{figure*}
Anomaly localization has brought a great influence on the intelligent industrial manufacturing and  extremely high valuable on extensive applications. It reveals the location and degree of various anomalies that are different from the normals. Due to the anomalies are unexpected and inexhaustible, it is challenging to construct a robust anomaly detector only with very limited normal data. Even though anomaly localization has spawned many superior unsupervised methods, the task is not yet completely resolved.

To handle with the shortage of real anomaly data, many appealing unsupervised approaches \cite{wacv21g2d, cutpaste, draem, jnld, omnial} have arose in accompany with anomaly synthesis. As examples illustrated in Fig.\ref{fig:1}c-e, existing synthesis methods \cite{cutpaste, draem, omnial} can only randomly destroy the appearance of normals shown in Fig.\ref{fig:1}a\cite{mvtecloco} and produce the anomalies having different visual structures with the normals. Fig.\ref{fig:1}b shows some more challenging cases that are beyond the capabilities of these synthesis methods. Each component in Fig.\ref{fig:1}b is flawless but their combination violates the certain underlying logical or geometrical constraints of the normal data. The anomaly detectors trained only with the synthetic structural anomalies would be vulnerable to logical anomalies. 

By comparing the difference between real logic anomaly (Fig.\ref{fig:1}b) and the synthetic anomaly (Fig.\ref{fig:1}d-e), we can see that existing synthesis methods \cite{draem, omnial} only modify discrete local regions and ignore the long-range dependencies. It leads to a more structure-oriented defects generation. Moreover, the fake boundaries introduced by a naive large region replacement make the anomaly detector learns a simple decision. Therefore, a photo-realistic logical anomaly synthesis method for unsupervised anomaly localization is highly demanded.

\begin{figure}
  \centering
  \setlength{\abovecaptionskip}{0.cm}
    \includegraphics[width=\linewidth]{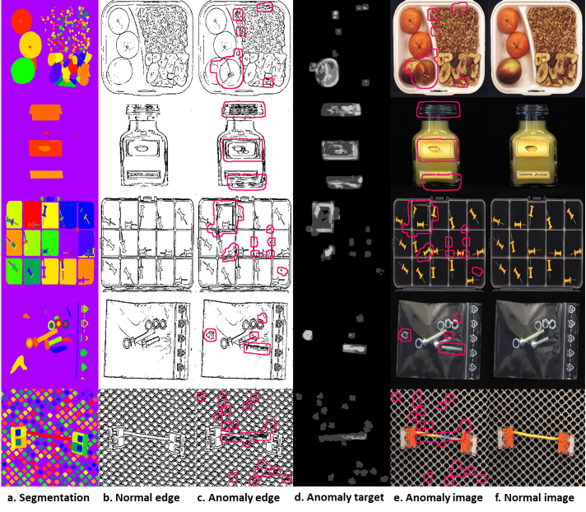}
  \caption{\textbf{Visualization of our synthetic anomalies.} (a) and (b) are extracted from (f). (c) is generated by merging the normal edges (b) from the selected regions of (a) with sampled edges. Anomaly image (e) is converted from (c). (d) is the difference map between (e) and (f).}
  \label{fig:3}
  \vspace{-1mm}
\end{figure}

\begin{figure*}
  \centering
  \setlength{\abovecaptionskip}{0.cm}
    \includegraphics[width=\linewidth]{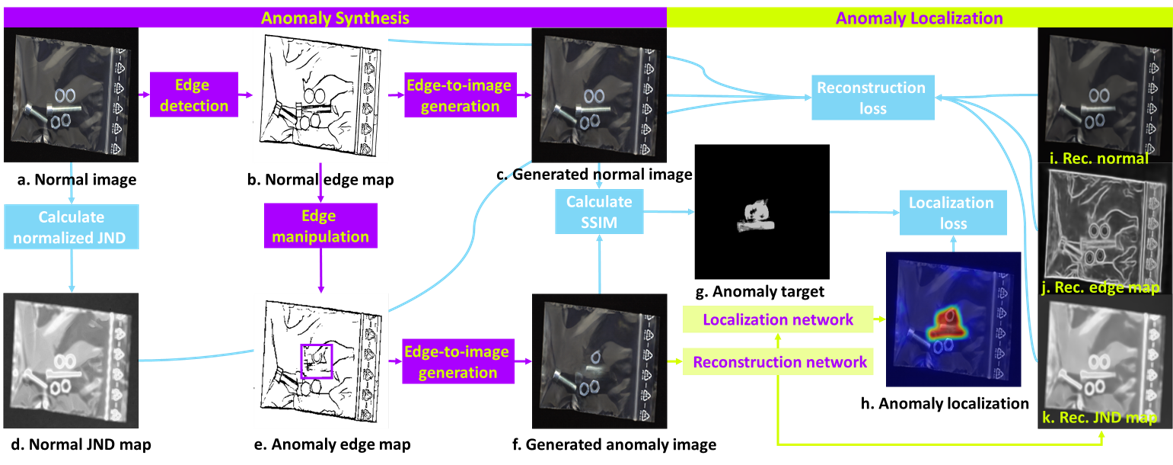}
  \caption{\textbf{Framework of proposed unsupervised anomaly localization.} It consists of anomaly synthesis and localization modules. Anomaly synthesis is based on anomaly edge map construction and edge-to-image generation. Synthetic anomaly is reconstructed into normal image, corresponding just noticeable distortion (JND) and edge maps. Anomaly localization is achieved by exploring difference between reconstructed and original data. }
  \label{fig:4}
  \vspace{-5mm}
\end{figure*}

To conquer aforementioned problems, we propose a novel anomaly synthesis and localization method LogicAL for effectively localizing anomaly pixels of both logical and structural anomalies. Fig.\ref{fig:2} illustrates overall concept of LogicAL. To generate photo-realistic anomalies, it manipulates edges and converts the modified edge map into image by using edge-to-image generator. To generate both logical and structural anomalies as shown in Fig.\ref{fig:2}g and Fig.\ref{fig:2}k, LogicAL modifies the normal edge map by using different region selection and edge manipulation strategies. By using pre-trained segmentation network SAM \cite{sam}, LogicAL can manipulate the edges from a semantic region. As shown in Fig.\ref{fig:2}c to Fig.\ref{fig:2}g, it syntheses a logical anomaly status that the breakfast-box is missing one orange. Fig.\ref{fig:2}h to Fig.\ref{fig:2}k demonstrate the structural anomaly generation. To forge structural anomaly edges, LogicAL replaces the normal edges from arbitrary regions with the augmented (flip or resize) edges that are sampled from the trainset. Beside removing and replacing normal edges, LogicAL also uses strategy of merging edges to syntheses more complex situations of mixing logical and structural anomalies, as shown in Fig.\ref{fig:3}. Fig.\ref{fig:4} shows the anomaly localization framework that equipped with LogicAL anomaly synthesis. It produces high quality reconstruction and precise anomaly localization for both logical and structural anomalies. 

In summary, we make following main contributions:
\begin{itemize} 
  \item We propose a logical anomaly synthesis method that generates photo-realistic anomaly images violating underlying logical constraints. It fills the gap left by existing structure-oriented anomaly synthesis methods.
  \item We introduce different edge manipulation and region selection strategies to balance logical and structural anomaly synthesis. By using the proposed anomaly synthesis, we improve existing reconstruction-based methods without effecting their inference runtime.   
  \item We further improve the anomaly localization performance by introducing edge reconstruction into the network structure. We achieve superior anomaly localization performance, 69.7 pixel AU-sPRO and 98.3 pixel AUROC, on challenge MVTecLOCO \cite{mvtecloco} and MVTecAD \cite{mvtec} datasets respectively.
\end{itemize}


\section{Related works}
\label{sec:relatedworks}

\textbf{Anomaly synthesis.} Anomaly synthesis is closely paired with unsupervised anomaly localization methods \cite{cutpaste, draem, jnld, pseudoad, dra, dsr} due to the lack of real anomaly samples. CutPaste \cite{cutpaste} propose a simple but effective method that syntheses anomaly by cutting a local rectangular region from the normal image and paste it back at a random position. It learns representations by classifying normal data from the synthetic anomalies. SPD \cite{visa} uses a smoothed version of CutPaste \cite{cutpaste} augmentation. Extensive experiments from Draem \cite{draem} show that the quality of synthetic anomaly highly effects the performance. Instead of using simple regular shaped anomalous, Draem \cite{draem} carefully designs the anomaly regions mask by using binarized Perlin noise. To increase the anomaly diversity, it extracts the anomaly texture from an extra DTD \cite{dtd} dataset. To simulate photo-realistic anomaly samples, JNLD \cite{jnld} proposes a multi-scale noticeable anomalous generation method based on just noticeable distortion \cite{jnd}. OmniAL \cite{omnial} further improves the anomaly synthesis of JNLD \cite{jnld} by using a panel-guided strategy to train a unified detector for N classes. Different with our method, these existing methods pay more attention on structural anomaly while ignore the logical anomaly.

To syntheses logical anomaly, SLSG \cite{slsg} extends the anomaly synthesis of Draem \cite{draem} by controlling the parameters of Perlin noise and the binarization threshold to get a more concentrated anomaly region mask. More recent road anomaly detection methods \cite{xia2020synthesize, lis2019detecting} re-synthesize the input image from the predicted semantic map using a generative adversarial network. Synboost \cite{di2021pixel} presents a pixel-wise anomaly detection framework that uses uncertainty maps to improve over existing re-synthesis methods \cite{xia2020synthesize, lis2019detecting} in finding dissimilarities between the input and generated images. However, these synthesis anomalies in road scene often introduce large difference in the normal regions. Inspired by existing methods, we propose a novel edge-controlled anomaly synthesis method that generates photo-realistic both logical and structural anomalies.

\textbf{Anomaly localization.} Anomaly localization aims to segment out the pixel-level anomaly regions. On the other hand, anomaly detection refers to distinguishing anomalous images at the image-level from the majority of normal images. Existing methods solve these two tasks either by a distance-based \cite{cutpaste, us, spade,padim,patchcore} or reconstruction-based \cite{us,draem,jnld,omnial,slsg} way.
The distance-based methods usually rely on feature extractors pre-trained on the ImageNet. CutPaste \cite{cutpaste} uses GradCAM \cite{gradcam} to get pixel-level anomaly location. SPADE \cite{spade} relies on K nearest neighbors of pixel-level feature pyramids extracted by pre-trained deep features. It detects anomaly based on alignment between an anomalous image and a constant number of the similar normal images.  Instead of using time-costly clustering, PaDim \cite{padim} uses a well-known Mahalanobis distance metric \cite{mahalanobis1936generalized} as an anomaly score. PatchCore \cite{patchcore} combines patch-level embeddings from ImageNet models with an outlier detection model. The reconstruction-based methods \cite{draem, jnld, omnial, slsg} use segmentation sub-network to predict the defective regions by comparing the difference between reconstructed images with input. 

Student-teacher (S-T) \cite{us} network uses three different pre-trained CNN as teachers with different receptive fields. It trains three student networks to mimic their corresponding teachers on the normal images. The anomaly is revealed by the failure mimic on unseen anomalous images. To better detect both logical and structural anomalies, GCAD \cite{mvtecloco} introduces the global and local branches using S-T networks \cite{us}. The final prediction is obtained by merging results of the two branches. SLSG \cite{slsg} uses a generative pre-trained network to learn the feature embedding of normal images. To model position information in images, it uses a self-supervised task to learn the reasoning of position relationships and use the graph convolutional network to capture across-neighborhood position relationships. This paper differs from these previous works by learning edge information from the edge-controlled synthetic anomaly.

\section{Methods}
\label{sec:methods}

As shown in Fig.\ref{fig:4}, the proposed method LogicAL consists of anomaly synthesis and anomaly localization. The anomaly synthesis module yields photo-realistic anomalies by globally and locally edge control. It extracts edge with a pre-trained edge detector and converts the modified edge into anomaly image by a pre-trained generator. During training, the anomaly synthesis module alternatively provides logical and structural anomalies to the anomaly localization module in real-time. The anomaly localization module learns to reconstruct various normal information and spot the difference between the reconstructed and original data. In testing phase, it only takes the anomaly localization module to derive the pixel-wise prediction.

\subsection{Anomaly synthesis}
\textbf{Edge detection.} To generate photo-realistic anomalies, it is better to tune an intermediate representation rather than directly modify the source image. Edge reveals sharp brightness, color or texture changes in any part of the image. Due to its unique object-agnostic property, edge is a suitable option to be the intermediate representation. With the development of deep learning, edge detection changes from the traditional gradient-based method to the end-to-end learning-based method. PiDiNet \cite{pidinet} is one of the appealing edge detectors that achieve a better trade-off between accuracy and efficiency. It integrates the advantages of traditional edge detectors and deep CNNs by using the pixel difference convolution. By using PiDiNet \cite{pidinet}, we achieve zero-shot edge detection for following edge control based anomaly synthesis. As shown in Fig.\ref{fig:4}, the extracted useful edge map Fig.\ref{fig:4}b is also used in training the anomaly localization module to avoid blurry reconstruction and inaccurate localization.

\begin{figure}
  \centering
  \setlength{\abovecaptionskip}{0.cm}
    \includegraphics[width=\linewidth]{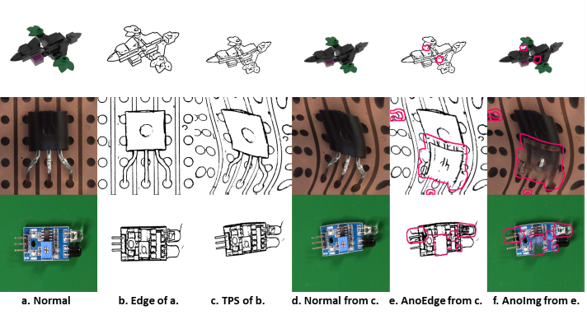}
  \caption{\textbf{Edge-to-image generation.} (a) Normal images from MADsim \cite{pad}, MVTecAD \cite{mvtec} and VisA \cite{visa}, (b) Edge maps extracted by PiDiNet \cite{pidinet}, (c) TPS \cite{tps} warping on (b), (d) Images generated from (c) by DeepSIM \cite{deepsim}, (e) and (f) are synthesis anomaly edge maps and images applied same TPS warping.}
  \label{fig:5}
  \vspace{-5mm}
\end{figure}

\textbf{Edge-to-image generation.} As shown in Fig.\ref{fig:5}, our edge-to-image generation module converts the extracted edge maps to color images with pixel-to-pixel correspondence. It is trained as the pix2pixHD \cite{pix2pixhd}, a conditional generative adversarial network (cGAN), with pairs of edge maps and color images from the normal dataset. The pix2pixHD \cite{pix2pixhd} consists of a coarse-to-fine generator translating edge maps to color images and a multi-scale discriminator distinguishing real images from the generated ones. It aims to model the conditional distribution of real images given the input edge maps via the minimax game. Considering the need of generation from anomaly edge, we follow DeepSIM \cite{deepsim} to augment the training pairs by using thin-plate-spline (TPS) \cite{tps} warps. The TPS augmentation simultaneously manipulates both edge map and color image by randomly shifting 3x3 control points in the horizontal and vertical directions. The warp is smoothed by further optimization \cite{tps}. Fig.\ref{fig:5}c-d show examples of TPS warped edge map and corresponding color image. By training with these smooth warped pairs, the generator is less likely to produce a global collapsed result when it handles the synthetic or real anomaly edges. 

\textbf{Edge manipulation.} Given the pre-trained edge-to-image generator, it is supposed to generate anomaly images from the  modified edge maps. Even though the generator is trained with carefully augmented data, it still produces global collapsed results if the input edge maps are out-of-distribution. Supplementary demonstrates that a suitable portion of anomaly edges is the key to generate high quality anomaly images. Therefore, our edge manipulation involves two region selection and three edge modification strategies. 

As illustrated in Fig.\ref{fig:2}, the region selection strategies consist of semantic and semantic-agnostic regions selection. We use the pre-trained segmentation model SAM \cite{sam} to get rough semantic regions. Due to SAM \cite{sam} is over-segmented, we further refine the segmentations by removing background, grouping small regions and merging overlapped regions and build a map of candidate regions, as shown in Fig.\ref{fig:2}d. For the semantic-agnostic regions selection, we randomly combine three regions with different aspect ratios or shapes.
  
Given the selected regions, we modify the normal edges by removing, replacing or merging strategies. The candidate anomaly edges are derived from the normal edges belong to the same class. By doing these edge modifications, it is easier to yield the logical anomalies that violate the constraints of amount, position, matching, etc. Fig.\ref{fig:3} shows examples of synthetic anomalies for MVTecLOCO \cite{mvtecloco} dataset. As the ambiguity exists between structural and logical anomalies, different strategies are not always strictly correspond to either logical or structural anomalies. These modifications also can fake the structural anomalies, such as scratches, dents, or contaminations.


Given the j-th normal image $I_j$, the corresponding anomaly image $I_{Aj}$ is generated as follow:
\begin{equation}
E_{Aj}=\left\{
\begin{aligned}
&M_j+(1-M_j)*E_j,remove \\
&M_j*A(E_{i})+(1-M_j)*E_j,replace \\
&M_j*(A(E_{i})+E_j-1)+(1-M_j)*E_j,merge
\end{aligned}
\right.
\end{equation}


where $E_j$ is the source normal edge map extracted from $I_j$, $E_i$ is the candidate anomaly edge map extracted from the i-th normal image $I_i$, $A$ indicates random augmentation applied to $E_i$, the source anomaly edge map $E_{j}$ is the combination of the augmented candidate anomaly edge maps $A(E_i)$, $M_i$ is the selected regions mask. As shown in Fig.\ref{fig:5}e and Fig.\ref{fig:5}f, the synthesis anomaly edge map and image can be further augmented by applying TPS \cite{tps} warping. 

\subsection{Anomaly localization}
Overall, our anomaly localization module follows the network structure of OmniAL \cite{omnial} that consists of reconstruction and localization sub-networks equipped with dilated channel and spatial attention (DCSA) blocks and DiffNeck module. The normalized just noticeable distortion (JND) map \cite{jnd, jnld, omnial} reveals a perceptual threshold of intensity change in an image that can be noticed by the human vision system. For both JND map and normal image reconstruction, we not only use MSE loss to supervise the pixel-to-pixel recovering but also the structural similarity (SSIM) \cite{ssim} loss to yield plausible local consistency. Different from OmniAL \cite{omnial}, we introduce edge information to the learning flow. In addition to learn to reconstruct the normal image and JND map, the reconstruction sub-network also produces the edge map. Following PiDiNet \cite{pidinet}, we adopt the annotator-robust loss function proposed in \cite{edgeloss} for the reconstructed edge maps. To localize anomaly regions more precisely, the localization sub-network calculate the reconstruction error with the assistance of edge maps. 
 
To train the localization sub-network, we can construct the ground truth maps by comparing the synthetic anomaly images with either the source normal images or the generated images from the normal edges. Due to the error of edge-to-image generator, the normal regions in the generated and the original images are not exactly same. It is impossible to localize anomaly region simply from the generated images. The SSIM \cite{ssim} provides a measure of the similarity by comparing two images based on luminance similarity, contrast similarity and structural similarity information. To avoid constructing an over-sensitive anomaly detector, we use SSIM to calculate the ground truth maps with both generated normal and anomaly images. To handle the unbalance of normal and anomaly, we use focal loss \cite{focalloss} to supervise the predicted anomaly localization.
\section{Experiments}
\label{sec:experiments}

To demonstrate the effectiveness of proposed LogicAL, we conduct extensive experiments on the challenging MVTecLOCO \cite{mvtecloco}, MVTecAD \cite{mvtec}, VisA \cite{visa} and MADsim \cite{pad} datasets. We evaluate the overall performance of anomaly detection and localization comparing with existing advanced methods. Our experiments also include ablation studies on the effectiveness of each component. 

\begin{table*}[t]
\begin{center}
\setlength{\abovecaptionskip}{0.cm}
\caption{Performance comparison of LogicAL with existing methods on MVTecLOCO\cite{mvtecloco}.} \label{table:1}
\resizebox{0.9\textwidth}{20mm}{
	\begin{tabular}{c|cccc|cccc}
	

  \hline
  \multirow{3}{*}{\textbf{Category}} 
  &\multicolumn{4}{c|}{Methods \textbf{W/O} anomaly synthesis}
  &\multicolumn{4}{c}{Methods \textbf{With} anomaly synthesis}
  
  \\
  \cline{2-9}
  & \multicolumn{1}{c|}{\textbf{Patchcore\cite{patchcore}}}
  & \multicolumn{1}{c|}{\textbf{SINBAD\cite{SINBAD}}}
  & \multicolumn{1}{c|}{\textbf{GCAD\cite{mvtecloco}}}
  & \multicolumn{1}{c|}{\textbf{EfficientAD\cite{efficientad}}}

  & \multicolumn{1}{c|}{\textbf{Draem\cite{draem}}} 
  & \multicolumn{1}{c|}{\textbf{SLSG\cite{slsg}}  }  
  & \multicolumn{1}{c|}{\textbf{OmniAL\cite{omnial}}}  
  & \multicolumn{1}{c}{\textbf{LogicAL}}\\
    \cline{2-9} & \multicolumn{8}{c}{\textcolor{gray}{Image-level detection AU-ROC}   /   Pixel-level localization AU-sPRO (FPR 5\%)}\\    
    
		\hline	
  		BreakfastBox 
  		& \textcolor{gray}{81.3} / 46.0
  		& \textcolor{gray}{92.0} / -
  		& \textcolor{gray}{83.9} / 50.2
  		& \textcolor{gray}{-} / -
  		
  		& \textcolor{gray}{80.3} / -  		  		
  		& \textcolor{gray}{\textbf{88.9}} / \textbf{65.9}
  		& \textcolor{gray}{75.9} / 46.5
  		& \textcolor{gray}{85.4} / 46.8
  		 \\
  		JuiceBottle 
  		& \textcolor{gray}{95.6} / 71.0
  		& \textcolor{gray}{94.9} / -
  		& \textcolor{gray}{99.4} / 91.0
  		& \textcolor{gray}{-} / -
  		
  		& \textcolor{gray}{94.3} / -

  		& \textcolor{gray}{99.1} / 82.0
  		& \textcolor{gray}{\textbf{99.5}} / 87.7
  		& \textcolor{gray}{98.5} / \textbf{91.3}
  		\\
  		Pushpins
  		& \textcolor{gray}{72.3} / 44.7 
  		& \textcolor{gray}{78.8} / -
  		& \textcolor{gray}{86.2} / 73.9
  		& \textcolor{gray}{-} / -
  		
  		& \textcolor{gray}{68.6} / -

  		& \textcolor{gray}{\textbf{95.5}} / 74.4
  		& \textcolor{gray}{79.6} / 59.6
  		& \textcolor{gray}{87.4} / \textbf{81.3}
  		 \\
  		ScrewBag 
  		& \textcolor{gray}{64.9} / 52.2
  		& \textcolor{gray}{85.4} / -
  		& \textcolor{gray}{63.2} / 55.8
  		& \textcolor{gray}{-} / -
  		
  		& \textcolor{gray}{70.6} / -

  		& \textcolor{gray}{79.4} / 47.2
  		& \textcolor{gray}{\textbf{83.1}} / 53.2
  		& \textcolor{gray}{82.0} / 52.3
  		\\
  		SConnector 
  		& \textcolor{gray}{82.4} / 58.6
  		& \textcolor{gray}{92.0} / -
  		& \textcolor{gray}{89.3} / 79.8
  		& \textcolor{gray}{-} / -
  		
  		& \textcolor{gray}{85.4} / -

  		& \textcolor{gray}{88.5} / 66.9
  		& \textcolor{gray}{88.1} / 69.1
  		& \textcolor{gray}{\textbf{89.0}} / \textbf{76.3}
  		\\  		
  		\hline	
  		Mean
  		& \textcolor{gray}{79.3} / 54.5
  		& \textcolor{gray}{86.8} / -
  		& \textcolor{gray}{83.3} / 70.1
  		& \textcolor{gray}{90.7} / 79.8
  		
  		& \textcolor{gray}{79.8} / -

  		& \textcolor{gray}{\textbf{90.3}} / 67.3
  		& \textcolor{gray}{85.0} / 63.0
  		& \textcolor{gray}{88.5} / \textbf{69.7}
  		\\
		\hline
\end{tabular}
}
\vspace{-0.5cm}
\end{center}
\end{table*}

\begin{table}[t]
\begin{center}
\setlength{\abovecaptionskip}{0.cm}
\caption{Performance comparison for both structural
and logical anomalies on MVTecLOCO\cite{mvtecloco}.} \label{table:2}
\resizebox{0.5\textwidth}{15mm}{
	\begin{tabular}{c|c|c|c}
	\hline
    \multirow{2}{*}{\textbf{Method}} 
  &  \textbf{Logical}
  &  \textbf{Structural}
  &  \textbf{Mean}
  \\
    \cline{2-4} & \multicolumn{3}{c}{\textcolor{gray}{Image AU-ROC}   /   Pixel AU-sPRO (FPR 5\%)}\\
		\hline	
  		\textbf{S-T\cite{us}}
  		& \textcolor{gray}{66.4}/49.7
  		& \textcolor{gray}{88.3}/75.6
  		& \textcolor{gray}{77.3}/62.6
  		 \\
  		\textbf{SPADE\cite{spade}} 
  		& \textcolor{gray}{\textbf{70.9}}/53.6
  		& \textcolor{gray}{66.8}/36.8
  		& \textcolor{gray}{68.9}/45.1
  		\\
  		\textbf{GCAD\cite{mvtecloco}} 
  		& \textcolor{gray}{86.0}/\textbf{71.1}
  		& \textcolor{gray}{80.6}/69.2
  		& \textcolor{gray}{83.3}/\textbf{70.1}
  		 \\
  		 \textbf{SLSG\cite{slsg}} 
  		& \textcolor{gray}{89.6}/-
  		& \textcolor{gray}{91.4}/-
  		& \textcolor{gray}{\textbf{90.3}}/67.3
  		 \\
  		\textbf{LogicAL}
  		& \textcolor{gray}{84.6}/68.8 
  		& \textcolor{gray}{\textbf{93.6}}/\textbf{81.5}
  		& \textcolor{gray}{88.5}/69.7
  		
  		\\
		\hline
\end{tabular}
}
\vspace{-0.5cm}
\end{center}
\end{table}

\begin{figure*}
  \centering
  \setlength{\abovecaptionskip}{0.cm}
    \includegraphics[width=\linewidth]{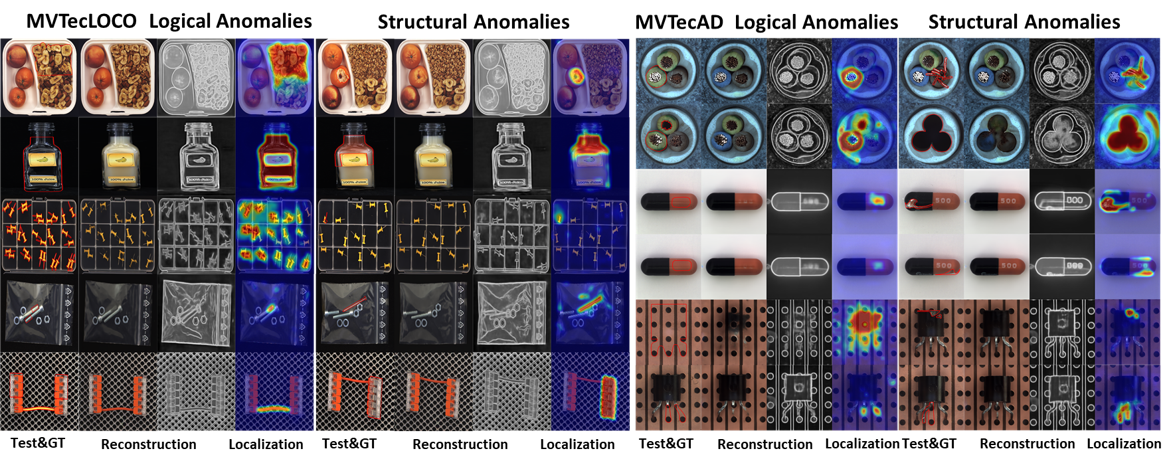}
  \caption{\textbf{Qualitative illustration of our anomaly detection results.} From left to right, the test images overlap with red ground truth contours, the reconstructions and the anomaly localization map overlays are shown in columns. The logical and structural anomalies of MVTecLOCO \cite{mvtecloco} and MVTecAD \cite{mvtec} are shown in groups.}
  \label{fig:6}
  \vspace{-0.3cm}
\end{figure*}

\begin{table}[t]
\begin{center}
\setlength{\abovecaptionskip}{0.cm}
\caption{Ablation study on MVTecLOCO \cite{mvtecloco}. Semantic: regions are extracted from refined SAM \cite{sam} segmentation, Arbitrary: random regions, Remove/Replace/Merge edges in the selected regions, TPS: apply TPS warping to synthetic anomaly edge maps and images, I: Image-level classification AUROC, P: Pixel-level localization AUsPRO.} \label{table:3}
\resizebox{0.5\textwidth}{15mm}{
\begin{tabular}{c|c|c|c|c|c|c|c}
	\hline
 
  \multicolumn{2}{c|}{\textbf{Region selection}} 
  & \multicolumn{3}{c|}{\textbf{Edge modification}} 
  & \multicolumn{1}{c|}{\textbf{Aug}} 
  & \multicolumn{2}{c}{\textbf{Performance}} 
  \\
   \hline
  Semantic
  &Arbitrary 
  &Remove
  &Replace 
  &Merge
  &TPS
  &I
  &P
   \\
   \hline 
  	-
  	& - 
	& -  	
  	& -  
  	& -  	  
  	& - 	
  	& 85.2  
  	& 63.2 
\\
  	\hline 
  	- 
  	& $\surd$
	& -  	
  	& $\surd$     
  	& -	
  	& -   	
  	& 87.6 
  	& 67.4 
 \\ 
 \hline 
  	$\surd$
  	& - 
	& -  	
  	& $\surd$  
  	& -  	  
  	& - 	
  	& 89.0  
  	& 68.6 
\\
\hline 
  	$\surd$
  	& - 
	& $\surd$  	
  	& -  
  	& -  	  
  	& - 	
  	& 89.1  
  	& 68.1 
\\

\hline 
  	$\surd$
  	& - 
	& -  	
  	& -  
  	& $\surd$  	  
  	& - 	
  	& 88.5  
  	& 67.8 
\\
  \hline 
  	$\surd$
  	& - 
	& $\surd$  	
  	& $\surd$  
  	& -  	  
  	& - 	
  	& 88.7  
  	& 67.9 
\\
	
  	\hline 
  	$\surd$ 
	& -   	   	  	
  	& $\surd$ 
  	& $\surd$ 
  	& $\surd$ 
  	& - 
  	& 88.5  
  	& 67.0  
\\
  	
  \hline 
  	$\surd$ 	
  	& $\surd$    	  	
  	& $\surd$
  	& $\surd$ 
  	& $\surd$
  	& - 
  	& 88.5
  	& \textbf{69.7} 
\\
  	  \hline 
  	$\surd$
  	& $\surd$ 	   	  	
  	& $\surd$  
  	& $\surd$ 
  	& $\surd$
  	& $\surd$ 
  	& \textbf{89.2}  
  	& 67.2
\\  	  	  	
  \hline

\end{tabular}
}
\vspace{-0.5cm}
\end{center}
\end{table}

\begin{table*}[t]
\begin{center}
\setlength{\abovecaptionskip}{0.cm}
\caption{Performance comparison of LogicAL with existing methods on MVTecAD \cite{mvtec}. OmniAL+ indicates that the OmniAL network is trained with our synthetic anomalies.} \label{table:4}
\resizebox{0.9\textwidth}{32mm}{
\begin{tabular}{c|cc|ccccc}

  \hline
  \multirow{3}{*}{\textbf{Category}} 
  &\multicolumn{2}{c|}{Methods \textbf{W/O} anomaly synthesis}
  &\multicolumn{5}{c}{Methods \textbf{With} anomaly synthesis}
  
  \\
  \cline{2-8}
  & \multicolumn{1}{c|}{\textbf{Patchcore\cite{patchcore}}}  
  & \multicolumn{1}{c|}{\textbf{EfficientAD\cite{efficientad}}}

  & \multicolumn{1}{c|}{\textbf{Draem\cite{draem}}} 
  & \multicolumn{1}{c|}{\textbf{SLSG\cite{slsg}}  }  
  & \multicolumn{1}{c|}{\textbf{OmniAL\cite{omnial}}}
  & \multicolumn{1}{c|}{\textbf{OmniAL+}}  
  & \multicolumn{1}{c}{\textbf{LogicAL}}\\
    \cline{2-8} & \multicolumn{7}{c}{\textcolor{gray}{Image AU-ROC} / Pixel AU-ROC}\\
    
    \hline
  bottle 
  		& \textcolor{gray}{100}/98.6

  		& \textcolor{gray}{-}/-   
  		& \textcolor{gray}{99.1}/\textbf{99.1}
  		& \textcolor{gray}{99.4}/\textbf{99.1}  
  		& \textcolor{gray}{99.4}/99.0
  		& \textcolor{gray}{\textbf{100}}/98.2
  		& \textcolor{gray}{\textbf{100}}/95.9
  		\\
  		
  cable 
		& \textcolor{gray}{99.5}/98.4

  		& \textcolor{gray}{-}/-
  		& \textcolor{gray}{94.7}/94.7    
  		& \textcolor{gray}{98.3}/97.4  
  		& \textcolor{gray}{97.6}/97.1
  		& \textcolor{gray}{97.6}/97.2
  		& \textcolor{gray}{\textbf{99.0}}/\textbf{97.6}
  		\\
  		
  capsule 
		& \textcolor{gray}{98.1}/98.8   		
 		 
  		& \textcolor{gray}{-}/- 
  		& \textcolor{gray}{\textbf{98.5}}/94.3
  		& \textcolor{gray}{95.5}/95.9 
  		& \textcolor{gray}{92.4}/92.2
  		& \textcolor{gray}{95.5}/97.2
  		& \textcolor{gray}{95.5}/98.2
  		\\
  		
  carpet 
		& \textcolor{gray}{98.7}/99.0  		
  		 
  		& \textcolor{gray}{-}/- 
  		& \textcolor{gray}{95.5}/95.5
  		& \textcolor{gray}{99.0}/96.0  
  		& \textcolor{gray}{99.6}/\textbf{99.6}
  		& \textcolor{gray}{97.4}/99.3
  		& \textcolor{gray}{98.5}/99.5
  		\\
  		
  grid 
		& \textcolor{gray}{98.2}/98.7    		
  	   & \textcolor{gray}{-}/- 
  	   & \textcolor{gray}{99.9}/\textbf{99.7} 
  	   & \textcolor{gray}{\textbf{100}}/98.5 
  	   & \textcolor{gray}{\textbf{100}}/99.6
  	   & \textcolor{gray}{99.7}/99.5
  	   & \textcolor{gray}{99.4}/99.4
  	   \\
  	   
  hazelnut 
		& \textcolor{gray}{100}/98.7    		
   		   		  		  
  		& \textcolor{gray}{-}/-  
  		& \textcolor{gray}{\textbf{100}}/\textbf{99.7} 
  		& \textcolor{gray}{99.5}/97.8 
  		& \textcolor{gray}{98.0}/98.6
  		& \textcolor{gray}{99.9}/98.9
  		& \textcolor{gray}{98.5}/98.6
  		\\
  		
  leather 
		& \textcolor{gray}{100}/ 99.3   		
 		   		  		
  		& \textcolor{gray}{-}/- 
  		& \textcolor{gray}{\textbf{100}}/98.6
  		& \textcolor{gray}{\textbf{100}}/99.5 
  		& \textcolor{gray}{97.6}/99.7
  		& \textcolor{gray}{\textbf{100}}/\textbf{99.8}
  		& \textcolor{gray}{98.1}/99.5
  		\\
  		
  metal nut 
		& \textcolor{gray}{100}/ 98.4  		
  		 
  		& \textcolor{gray}{-}/-  
  		& \textcolor{gray}{98.7}/\textbf{99.5}
  		& \textcolor{gray}{\textbf{100}}/98.9  
  		& \textcolor{gray}{99.9}/99.1
  		& \textcolor{gray}{98.7}/96.6
  		& \textcolor{gray}{99.4}/97.6
  		\\
  		
  pill 
	& \textcolor{gray}{96.6}/97.4    	
  	   	 	
  	& \textcolor{gray}{-}/-  
  	& \textcolor{gray}{98.9}/97.6  
  	& \textcolor{gray}{\textbf{99.2}}/98.0 
  	& \textcolor{gray}{97.7}/98.6
  	& \textcolor{gray}{97.4}/98.5
  	& \textcolor{gray}{97.5}/99.1
  	\\
  	
  screw 
	& \textcolor{gray}{98.1}/99.4  	
 	 
  	& \textcolor{gray}{-}/-
  	& \textcolor{gray}{93.9}/97.6 
  	& \textcolor{gray}{89.1}/97.3  
  	& \textcolor{gray}{81.0}/97.2
  	& \textcolor{gray}{\textbf{100}}/98.9
  	& \textcolor{gray}{99.3}/99.3
  	\\
  	 
  tile 
	& \textcolor{gray}{98.7}/95.6    	

  	& \textcolor{gray}{-}/-  
  	& \textcolor{gray}{99.6}/99.2 
  	& \textcolor{gray}{\textbf{100}}/98.6  
  	& \textcolor{gray}{\textbf{100}}/99.4
  	& \textcolor{gray}{\textbf{100}}/98.4
  	& \textcolor{gray}{\textbf{100}}/99.0
  	\\
  	
  toothbrush 
	& \textcolor{gray}{100}/98.7    	

  	& \textcolor{gray}{-}/-   
  	& \textcolor{gray}{\textbf{100}}/98.1 
  	& \textcolor{gray}{\textbf{100}}/99.4 
  	& \textcolor{gray}{\textbf{100}}/99.2
  	& \textcolor{gray}{99.7}/\textbf{99.6}
  	& \textcolor{gray}{\textbf{100}}/99.4
  	\\
  	
  transistor 
	& \textcolor{gray}{100}/96.3    	
 	
  	& \textcolor{gray}{-}/- 
  	& \textcolor{gray}{93.1}/90.9
  	& \textcolor{gray}{97.3}/92.5  
  	& \textcolor{gray}{93.8}/91.7
  	& \textcolor{gray}{95.0}/91.0
  	& \textcolor{gray}{\textbf{98.2}}/\textbf{97.3}
  	\\
  	
  wood 
	& \textcolor{gray}{99.2}/95.0  	
  	
  	& \textcolor{gray}{-}/-   
  	& \textcolor{gray}{99.1}/96.4 
  	& \textcolor{gray}{99.6}/96.8  
  	& \textcolor{gray}{98.7}/\textbf{96.9}
  	& \textcolor{gray}{94.1}/94.3
  	& \textcolor{gray}{99.0}/93.9
  	\\
  	
  zipper 
	& \textcolor{gray}{99.4}/98.8    	

  	& \textcolor{gray}{-}/-  
  	& \textcolor{gray}{\textbf{100}}/98.8 
  	& \textcolor{gray}{\textbf{100}}/97.1  
  	& \textcolor{gray}{\textbf{100}}/\textbf{99.7}
  	& \textcolor{gray}{\textbf{100}}/\textbf{99.7}
  	& \textcolor{gray}{\textbf{100}}/99.5
  	\\
  	
  \hline
  Mean 
	& \textcolor{gray}{99.1}/98.1   	
  	
  	& \textcolor{gray}{99.1}/-  
  	& \textcolor{gray}{98.0}/97.3 
  	& \textcolor{gray}{98.5}/97.5  
  	& \textcolor{gray}{97.0}/97.8
  	& \textcolor{gray}{98.3}/97.8
  	& \textcolor{gray}{\textbf{98.8}}/\textbf{98.3}
  	\\
  \hline
\end{tabular}
}
\vspace{-0.5cm}
\end{center}
\end{table*}

\begin{table*}[t]
\begin{center}
\setlength{\abovecaptionskip}{0.cm}
\caption{Performance comparison of LogicAL with existing methods on VisA\cite{visa}. OmniAL+ indicates that the OmniAL network is trained with our synthetic anomalies. I-AUC: Image AUROC, P-AUC: Pixel AUROC, P-AP: Pixel AP.} \label{table:5}
\resizebox{0.9\textwidth}{20mm}{
\begin{tabular}{c|c|cc|ccc|ccc|ccc|ccc|ccc}
	\hline
 \multirow{2}{*}{}
  &\multirow{2}{*}{\textbf{Category}} 
  & \multicolumn{2}{c|}{\textbf{Padim+SPD\cite{visa}}} 
  & \multicolumn{3}{c|}{\textbf{Draem\cite{draem}}} 
  & \multicolumn{3}{c|}{\textbf{JNLD\cite{jnld}}} 
  & \multicolumn{3}{c|}{\textbf{OmniAL\cite{omnial}}  }  
  & \multicolumn{3}{c|}{\textbf{OmniAL+}  }  
  & \multicolumn{3}{c}{\textbf{LogicAL}  }  \\
  \cline{3-19}
     &&I-AUC&P-AUC &I-AUC&P-AUC&P-AP& I-AUC&P-AUC&P-AP& I-AUC&P-AUC&P-AP & I-AUC&P-AUC&P-AP & I-AUC&P-AUC&P-AP\\
    
    \hline
 	\multirow{4}{*}{\textbf{Complex structure}} 
  		& PCB1 
  		& 92.7 & 97.7 
  		& 71.3 & 98.6 & 60.4
  		& 82.0 & 96.4 & \textbf{72.8}
  		& 96.6 & 98.7 & 63.5 
  		& \textbf{97.3} & \textbf{99.2} & 62.4 
  		& 95.0 & 99.0 & 72.1 \\

  		& PCB2 
  		& 87.9 & \textbf{97.2}
  		& 89.7 & 92.5 & 3.5
  		& 96.3 & 91.9 & 31.2
  		& 99.4 & 83.2 & 2.8 
  		& \textbf{99.5} & \textbf{94.3} & \textbf{33.3}
  		& 97.8 & 93.9 & 31.4\\
  		
  		& PCB3 
  		& 85.4 & 96.7
  		& 73.1 & 93.8 & 18.7
  		& 96.9 & 95.3 & 43.4
  		& 96.9 & 98.4 & 56.9 
  		& \textbf{98.4} & \textbf{99.6} & \textbf{62.7} 
  		& 94.1 & 98.9 & 52.0 \\
  		
  		& PCB4 
  		& 99.1 & 89.2
  		& 91.3 & 95.8 & 32.7
  		& 94.8 & 96.1 & 37.4
  		& 97.4 & 98.5 & 38.4
  		& \textbf{98.0} & \textbf{98.6} & 49.0 
  		& 97.4 & \textbf{98.6} & \textbf{50.5} \\
  	\hline
  	\multirow{4}{*}{\textbf{Multiple instances}} 
  	
  		& Macaroni1 
  		& 85.7 & 98.8
  		& 70.3 & 95.8 & 8.2
  		& 94.3 & 98.8 & \textbf{25.9}
  		& \textbf{96.9} & 98.9 &7.6  
  		& 95.9 & 97.4 & 17.4 
  		& 94.0 & \textbf{99.0} & 8.5 \\
  		
  		& Macaroni2 
  		& 70.8 & 96.0
  		& 71.3 & 94.1 & \textbf{25.4}
  		& 86.5 & 92.9 & 17.2
  		& \textbf{89.9} & \textbf{99.1} & 11.4 
  		& 85.0 & 96.9 & 21.3
  		& 86.6 & \textbf{99.1} & 20.9 \\
  		
  		& Capsules 
  		& 68.1 & 86.3
  		& 77.3 & 93.7 & 20.2
  		& 89.1 & 98.9 & 38.6
  		& 87.9 & 98.6 & 62.9 
  		& 91.6 & 99.3 & 61.4 
  		& \textbf{93.9} & \textbf{99.7} & \textbf{65.1} \\
  		
  		& Candles 
  		& 89.1 & \textbf{97.3}
  		& 82.3 & 87.0 & 27.9
  		& 89.1 & 94.8 & 25.9
  		& 85.1 & 90.5 & \textbf{29.2} 
  		& 89.6 & 93.3 & 20.0 
  		& \textbf{92.4} & 90.2 &23.6 \\
  	\hline	
  	\multirow{4}{*}{\textbf{Single instance}} 
  		& Cashew 
  		& 90.5 & 86.1
  		& 94.2 & 94.7 & 41.2
  		& 96.0 & 96.3 & 43.7
  		& 97.1 & \textbf{98.9} & \textbf{77.3} 
  		& 97.3 & 97.6 & 81.1 
  		& \textbf{97.9} &97.6 &66.3 \\
  		
  		& Chewing gum 
  		& \textbf{99.3} & 96.9
  		& 93.4 & 97.5 &40.9
  		& 98.5 & 99.4 & 74.7
  		& 94.9 & 98.7 &\textbf{82.9} 
  		& 96.1 & 99.4 & 80.6
  		& 98.5 &\textbf{99.5} &73.8 \\
  		
  		& Fryum 
  		& 89.8 & 88.0
  		& \textbf{100} & \textbf{97.5} & 40.9
  		& 93.2 & 95.8 & \textbf{42.9}
  		& 97.0 & 89.3 & 28.3 
  		& 96.8 & 95.1 & 31.0 
  		& 99.0 &93.7 &29.2 \\
  		
  		& Pipe fryum 
  		& 95.6 & 95.4
  		& 94.1 & 81.8 & 23.7
  		& 96.0 & 97.0 & 44.1
  		& 91.4 & 99.1 & 69.1 
  		& \textbf{97.1} & 98.6 & 70.4 
  		& 97.0 & \textbf{99.5} & \textbf{72.9} \\
  	\hline	
  	& Mean
  		& 87.8 & 93.8
  		& 84.1 & 88.8 & 25.4
  		& 93.0 & 96.1 & 41.5
  		& 94.2 & 96.0 & 44.2
  		& 95.2 & \textbf{97.5} & \textbf{49.2} 
  		& \textbf{95.3} & 97.4 & 47.2 \\
\hline

\end{tabular}
}
\vspace{-0.5cm}
\end{center}
\end{table*}

\begin{table*}[t]
\begin{center}
\setlength{\abovecaptionskip}{0.cm}
\caption{Performance comparison of LogicAL with existing methods on MADsim \cite{pad}.} \label{table:6}
\resizebox{0.9\textwidth}{36mm}{
\begin{tabular}{c|cccc|ccc}
  \hline
  \multirow{3}{*}{\textbf{Category}} 
  &\multicolumn{4}{c|}{Methods \textbf{W/O} anomaly synthesis}
  &\multicolumn{3}{c}{Methods \textbf{With} anomaly synthesis}
  
  \\
  \cline{2-8}
  & \multicolumn{1}{c|}{\textbf{Patchcore\cite{patchcore}}}
  & \multicolumn{1}{c|}{\textbf{CFlow\cite{wacv22cflow}}}    
  & \multicolumn{1}{c|}{\textbf{UniAD\cite{uniad}}}  
  & \multicolumn{1}{c|}{\textbf{PAD\cite{pad}}} 
  & \multicolumn{1}{c|}{\textbf{Cutpaste\cite{cutpaste}}  }  
  & \multicolumn{1}{c|}{\textbf{Draem\cite{draem}}}  
  & \multicolumn{1}{c}{\textbf{LogicAL}}\\
    \cline{2-8} & \multicolumn{7}{c}{\textcolor{gray}{Image AU-ROC} / Pixel AU-ROC}\\
    
    \hline
  	Gorilla 
  		& \textcolor{gray}{66.8}/88.4 
  		& \textcolor{gray}{69.2}/94.7
  		& \textcolor{gray}{56.6}/93.4
  		& \textcolor{gray}{93.6}/99.5
  		& \textcolor{gray}{-}/36.1   
  		& \textcolor{gray}{\textbf{58.9}}/77.7 
  		& \textcolor{gray}{58.2}/\textbf{91.7}
  		\\
  		
  Unicorn 
		& \textcolor{gray}{92.4}/58.9   		 
  		& \textcolor{gray}{82.3}/89.9   
  		& \textcolor{gray}{73.0}/86.8
  		& \textcolor{gray}{94.0}/98.2
  		
  		& \textcolor{gray}{-}/69.6   
  		& \textcolor{gray}{70.4}/26.0
  		& \textcolor{gray}{\textbf{77.0}}/\textbf{82.4}
  		\\
  		
  Mallard 
		& \textcolor{gray}{59.3}/66.1   			  		
  		& \textcolor{gray}{74.9}/87.3 
  		& \textcolor{gray}{70.0}/85.4
  		& \textcolor{gray}{84.7}/97.4
  		
  		& \textcolor{gray}{-}/40.9 
  		& \textcolor{gray}{34.5}/\textbf{47.8} 
  		& \textcolor{gray}{\textbf{80.0}}/33.2
  		\\
  		
  Turtle 
		& \textcolor{gray}{87.0}/77.5  		  		
  		& \textcolor{gray}{51.0}/90.2 
  		& \textcolor{gray}{50.2}/88.9
  		& \textcolor{gray}{95.6}/99.1
  		
  		  		& \textcolor{gray}{-}/77.2 
  		& \textcolor{gray}{18.4}/45.3  
  		& \textcolor{gray}{\textbf{84.0}}/\textbf{90.1}
  		\\
  		
  Whale 
		& \textcolor{gray}{86.0}/60.9    		  	      	   
  	   & \textcolor{gray}{57.0}/89.2  
  	   & \textcolor{gray}{75.5}/90.7
  	   & \textcolor{gray}{82.5}/98.3
  	   
  	   & \textcolor{gray}{\textbf{-}}/66.8
  	   & \textcolor{gray}{65.8}/55.9 
  	   & \textcolor{gray}{\textbf{74.2}}/\textbf{86.7}
  	   \\
  	   
  Bird 
		& \textcolor{gray}{82.9}/88.6    		   		   		
  		& \textcolor{gray}{75.6}/91.8   
  		& \textcolor{gray}{74.7}/91.1
  		& \textcolor{gray}{92.4}/95.7
  		
  		& \textcolor{gray}{-}/71.7  
  		& \textcolor{gray}{69.1}/60.3 
  		& \textcolor{gray}{\textbf{71.7}}/\textbf{92.9}
  		\\
  		
  Owl 
		& \textcolor{gray}{72.9}/86.3   		 		   		
  		& \textcolor{gray}{76.5}/94.6
  		& \textcolor{gray}{65.3}/92.8
  		& \textcolor{gray}{88.2}/99.4
  		
  		& \textcolor{gray}{-}/51.9 
  		& \textcolor{gray}{\textbf{67.2}}/78.9 
  		& \textcolor{gray}{53.5}/\textbf{88.0}
  		\\
  		
  Sabertooth 
		& \textcolor{gray}{76.6}/69.4  		  		  		
  		& \textcolor{gray}{71.3}/93.3 
  		& \textcolor{gray}{61.2}/90.3
  		& \textcolor{gray}{98.5}/95.7
  		
  		  		& \textcolor{gray}{-}/71.2  
  		& \textcolor{gray}{68.6}/26.2  
  		& \textcolor{gray}{\textbf{70.2}}/\textbf{86.7}
  		\\
  		
  Swan 
	& \textcolor{gray}{75.2}/73.5    	  	   	
  	& \textcolor{gray}{67.4}/93.1  
  	& \textcolor{gray}{57.5}/90.6
  	& \textcolor{gray}{98.8}/86.5
  	
  	  	& \textcolor{gray}{-}/57.2  
  	& \textcolor{gray}{59.7}/75.9 
  	& \textcolor{gray}{\textbf{64.3}}/\textbf{91.7}
  	\\
  	
  Sheep 
	& \textcolor{gray}{89.4}/79.9  	  	  
  	& \textcolor{gray}{80.9}/94.3  
  	& \textcolor{gray}{70.4}/92.9
  	& \textcolor{gray}{90.1}/97.7
  	
  	  	& \textcolor{gray}{-}/67.2 
  	& \textcolor{gray}{59.5}/70.5  
  	& \textcolor{gray}{\textbf{79.5}}/\textbf{93.5}
  	\\
  	 
  Pig 
	& \textcolor{gray}{85.7}/83.5    	   	  	 
  	& \textcolor{gray}{72.1}/97.1 
  	& \textcolor{gray}{54.6}/94.8
  	& \textcolor{gray}{88.3}/97.7
  	
  	  	& \textcolor{gray}{-}/52.3  
  	& \textcolor{gray}{64.4}/65.6  
  	& \textcolor{gray}{\textbf{64.9}}/\textbf{92.4}
  	\\
  	
  Zalika 
	& \textcolor{gray}{68.2}/64.9    	   	   	 
  	& \textcolor{gray}{66.9}/89.4  
  	& \textcolor{gray}{50.5}/86.7
  	& \textcolor{gray}{88.2}/99.1
  	
  	  	& \textcolor{gray}{-}/43.5   
  	& \textcolor{gray}{51.7}/66.6 
  	& \textcolor{gray}{\textbf{52.9}}/\textbf{83.4}
  	\\
  	
  Phoenix 
	& \textcolor{gray}{71.4}/62.4

  	& \textcolor{gray}{64.4}/87.3
 
  	& \textcolor{gray}{93.8}/91.7
  	& \textcolor{gray}{82.3}/99.4
  	
  	  	& \textcolor{gray}{-}/53.1 
  	& \textcolor{gray}{53.1}/38.7 
  	& \textcolor{gray}{\textbf{59.7}}/\textbf{83.9}
  	\\
  	
  Elephant 
	& \textcolor{gray}{78.6}/56.2

  	& \textcolor{gray}{70.1}/72.4 

  	& \textcolor{gray}{55.4}/84.7
  	& \textcolor{gray}{92.5}/99.0
  	  	& \textcolor{gray}{-}/56.9   
  	& \textcolor{gray}{\textbf{62.5}}/55.9  
  	& \textcolor{gray}{57.0}/\textbf{82.3}
  	\\
  	
  Parrot 
	& \textcolor{gray}{78.0}/70.7

  	& \textcolor{gray}{67.9}/86.8 

  	& \textcolor{gray}{53.4}/85.6
  	& \textcolor{gray}{97.0}/99.5
  	  	& \textcolor{gray}{-}/55.4  
  	& \textcolor{gray}{\textbf{62.3}}/34.4  
  	& \textcolor{gray}{57.1}/\textbf{83.9}
  	\\
  
  Cat 
	& \textcolor{gray}{78.7}/85.6

  	& \textcolor{gray}{65.8}/94.7 

  	& \textcolor{gray}{53.1}/93.8
  	& \textcolor{gray}{84.9}/97.7
  	  	& \textcolor{gray}{-}/58.3  
  	& \textcolor{gray}{61.3}/79.4  
  	& \textcolor{gray}{\textbf{64.6}}/\textbf{95.4}
  	\\
  
    Scorpion 
	& \textcolor{gray}{82.1}/79.9

  	& \textcolor{gray}{79.5}/91.9 

  	& \textcolor{gray}{69.5}/92.2
  	& \textcolor{gray}{91.5}/95.9
  	  	& \textcolor{gray}{-}/71.2  
  	& \textcolor{gray}{\textbf{83.7}}/79.7  
  	& \textcolor{gray}{78.9}/\textbf{92.1}
  	\\
  	
  	Obesobeso 
	& \textcolor{gray}{89.5}/91.9    	
   	  	
  	& \textcolor{gray}{80.0}/95.8 

  	& \textcolor{gray}{67.7}/93.6
  	& \textcolor{gray}{97.1}/98.0
  	  	& \textcolor{gray}{-}/73.3  
  	& \textcolor{gray}{\textbf{73.9}}/89.2  
  	& \textcolor{gray}{71.2}/\textbf{95.5}
  	\\
  	
  	Bear 
	& \textcolor{gray}{84.2}/79.5    	
   	  	
  	& \textcolor{gray}{81.4}/92.2 

  	& \textcolor{gray}{65.1}/90.9
  	& \textcolor{gray}{98.8}/99.3
  	  	& \textcolor{gray}{-}/68.8  
  	& \textcolor{gray}{\textbf{76.1}}/39.2  
  	& \textcolor{gray}{70.9}/\textbf{88.2}
  	\\
  	
  	Puppy 
	& \textcolor{gray}{65.6}/73.3    	
   	  	
  	& \textcolor{gray}{71.4}/89.6 

  	& \textcolor{gray}{55.6}/87.2
  	& \textcolor{gray}{93.5}/98.8
  	  	& \textcolor{gray}{-}/43.2  
  	& \textcolor{gray}{57.4}/45.8  
  	& \textcolor{gray}{\textbf{61.7}}/\textbf{85.8}
  	\\
  \hline
  Mean 
	& \textcolor{gray}{78.5}/74.7   	
  	  	 
  	& \textcolor{gray}{71.3}/90.8 
  
  	& \textcolor{gray}{62.2}/89.1
  	& \textcolor{gray}{90.9}/97.8
  	  	& \textcolor{gray}{-}/59.3  
  	& \textcolor{gray}{60.9}/58.0
  	& \textcolor{gray}{\textbf{67.6}}/\textbf{86.0}
  	\\
  \hline
\end{tabular}
}
\vspace{-0.5cm}
\end{center}

\end{table*}
\subsection{Datasets}
\textbf{MVTecLOCO} \cite{mvtecloco} dataset contains five categories of approximately 3,644 images covering both logical and structural anomalies
from industrial inspection scenarios. 
Considering the ambiguity of the logical anomaly determination at the pixel level, MVTecLOCO \cite{mvtecloco} introduces a performance metric, saturated per-region overlap (sPRO), that takes the different modalities of the defects present in the dataset into account. It is a generalized version of the per-region overlap (PRO) metric that saturates once the overlap with the ground truth exceeds a certain saturation threshold. To evaluate pixel-level anomaly localization performance, we calculate the AU-sPRO score based on the area under the FPR-sPRO curve up to the false positive rate is 5\%. 
The performance of image-level classification is evaluated by the AU-ROC.


\textbf{MVTecAD} \cite{mvtec} dataset contains 1,258 test images but pays more attention on structural anomalies than MVTecLOCO \cite{mvtecloco}. It only includes 37 test images from three out of fifteen categories matching the definition of logical anomalies, including cable, capsule and transistor. As shown in Fig.\ref{fig:6}, the logical anomalies consists of cable swap, faulty imprint and transistor misplaced. Follow \cite{padim, draem, jnld, slsg, omnial}, we evaluate our method on anomaly detection and localization with image and pixel AU-ROC.  

\textbf{VisA} \cite{visa} dataset consists of 9,621 normal and 1,200 anomalous color images covering 12 objects in 3 domains, including complex structure, multiple instances and single instances. The anomalous images contain various flaws, including surface defects such as scratches, dents, color spots or crack, and logical defects like misplacement or missing parts. There are 5-20 images per defect type and an image may contain multiple defects. 

\textbf{MADsim} \cite{pad} dataset contains 5,231 normal and 4,902 anomaly color images of 20 types of 3D LEGO animal models from different viewpoints covering a wide range of poses. It uses Blender in combination with Ldrew (LEGO parts library) to build three types of anomalous, such as burrs, stains and missing parts. To evaluate pose-agnostic anomaly detection ability, we also evaluate our method on MADsim dataset. Fig.\ref{fig:5} illustrates examples from MADsim \cite{pad}, MVTecAD \cite{mvtec} and VisA \cite{visa} datasets.

\subsection{Implementation}
We extract four scales edge maps with pre-trained PiDiNet \citep{pidinet}. Since the first scale edge map contains more detail edges, we further use it to train edge-to-image generators and anomaly localization network. With the pairs of extracted edge map and color image, we firstly train an unified DeepSIM \citep{deepsim} model 300 epochs with a batch size of 56 images having size of 256x256 for each dataset. Different edge manipulation strategies are used to synthesis anomaly edge maps alternatively. The pre-trained DeepSIM models are used to convert synthesis anomaly edge maps to anomaly images on-the-fly during training the anomaly localization network. For anomaly localization, we train the network 300 epochs with a batch size of 15(\textcolor{gray}{12}) images having 
size of 256x256(\textcolor{gray}{256x320}) for MVTecAD \cite{mvtec}, MVTecLOCO \cite{mvtecloco} and MADsim \cite{pad}(\textcolor{gray}{VisA \cite{visa}}). The Adam optimizer has an initial learning rate of 1e-4 and decreases the learning rate with multi-step schedule. 

\subsection{Performance}
\textbf{Quantitative evaluation.} Table \ref{table:1} and \ref{table:2} show our quantitative comparison with the state-of-the-art methods recently reported on MVTecLOCO \cite{mvtecloco} for both logical and structural anomaly detection and localization. Performance of the other methods in Table \ref{table:1} are reported from SLSG \cite{slsg}. Separate performance comparison of logical and structural anomalies are shown in Table \ref{table:2}. Overall, we achieve better comprehensive performance of anomaly detection (88.5 AU-ROC) and localization (69.7 AU-sPRO) than existing methods. Even though SLSG \cite{slsg} achieves good performance in anomaly detection with 90.3 AU-ROC, it only has 67.3 AU-sPRO score in anomaly localization. The similar situation happens to GCAD \cite{mvtecloco} that is proposed along with the MVTecLOCO dataset. Table \ref{table:4} shows our superior performance comparing with anomaly synthesis based methods on the MVTecAD dataset. By training with our synthetic anomalies, we improve OmniAL \cite{omnial} by 1\% performance on image-level anomaly detection. With edge reconstruction, our method LogicAL achieves further 0.5\% improvement on both anomaly detection and localization. Table \ref{table:5} illustrates our performance on VisA \cite{visa} comparing with existing methods. We also achieve 1\% improvement for both image and pixel level from the baseline method. Table \ref{table:6} demonstrates our capability of multi-pose anomaly detection on the challenge MADsim \cite{pad}. Without using pose alignment process, we achieve superior performance than other anomaly synthesis based methods.

\textbf{Qualitative evaluation.} We visualize the pixel-wise prediction of our method on MVTecLOCO \cite{mvtecloco} and MVTecAD \cite{mvtec} in Fig.\ref{fig:6}. The reconstruction sub-network generates high quality normal version of the received anomaly input. The localization sub-network precisely reveals the anomaly regions by comparing the difference between the reconstructed images and input. For example, in the second row of the MVTecLOCO results, the anomalous empty bottle is filled with white juice according to the banana sticker. The empty regions excepted the stickers parts are predicted as anomalies. Similarly, the red juice doesn't belong to banana sticker is corrected into white juice and highlighted as anomalies. In the examples of capsule from MVTecAd dataset, the faulty imprint of '500' is recovered both in color image and edge map. The cracked and squeezed capsule are reconstructed into normal appearance. The corresponding anomalous regions are revealed. More results are shown in supplementary material.

\subsection{Ablation Study}
Table \ref{table:3} demonstrates the effectiveness of proposed edge manipulation based anomaly synthesis. We carry on ablation experiments on MVTecLOCO with different region selection and edge modification strategies and TPS augmentation. For region selection strategy, we compare the performance of using both or either semantic and arbitrary regions for edge modification. For edge modification strategy, we evaluate the effectiveness of removing, replacing and merging edges in the selected regions. We also evaluate the contribution of TPS warping augmentation on synthetic anomalies. Our baseline is OmniAL \cite{omnial} that achieves 85.2 image AUROC and 63.2 pixel AUsPRO. By training with the anomalies synthesised by modifying edges in semantic regions, we achieves more than 3\% improvement in both image-level detection and pixel-level localization. By combining all edge modification strategies, we improve the performance to 67.0 pixel AUsPRO. The arbitrary region selection strategy increase the diversity of training samples and brings 2.7\% improvement. By applying TPS warping augmentation, we achieve 0.7\% improvement for image AUROC but decrease 2.5\% pixel AUsPRO.


\section{Conclusion}
\label{sec:conclusion}

Different with existing methods, this paper proposes a novel anomaly synthesis method for both unsupervised logical and structural anomaly localization. It generates photo-realistic anomaly images violating underlying logical constraints and fills the gap left by existing structure-oriented anomaly synthesis methods. More specifically, it introduces edge manipulation strategies to balance logical and structural anomaly synthesis. By modifying edges in semantic regions, it easily generates anomalies that break logical constraints, such as missing components. By editing edges in arbitrary regions, it forges varies structural defects. It further improves the anomaly localization performance by introducing edge reconstruction into the network structure. Extensive experiments on the challenge dataset verify the advantages of proposed method.

{
    \small
    \bibliographystyle{ieeenat_fullname}
    \bibliography{main}
}
\clearpage
\setcounter{page}{1}
\maketitlesupplementary


\section{Ablation study}
\label{sec:ablationstudy}
\begin{figure}
  \centering
  \setlength{\abovecaptionskip}{0.cm}
    \includegraphics[width=\linewidth]{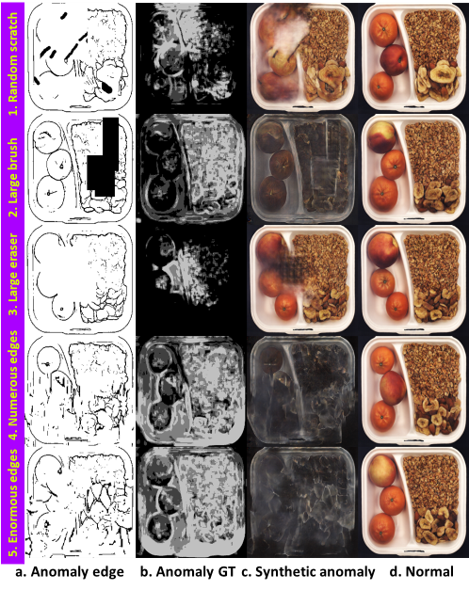}
  \caption{\textbf{Analysis of edge manipulation.} Improper anomaly edges cause collapse of edge-to-image generation.}
  \label{fig:7}
  \vspace{-5mm}
\end{figure}

\textbf{Edge manipulation.} Given the pre-trained edge-to-image generator, it is supposed to generate anomaly images from the  modified edge maps. Even though the generator is trained with carefully augmented data, it still produces global collapsed results if the input edge maps are out-of-distribution. Fig.\ref{fig:7} illustrates the failure generations given anomaly edge maps from 5 different modifications. Overall, the modifications are based on the concerns of adding and/or removing different scales of edges. As shown in Fig.\ref{fig:7}(2), the edges forged by large brush are obviously different from the real edges. Fig.\ref{fig:7}(4)-(5) show that scale matters. Either numerous nor enormous modified edges make anomaly synthesis successfully. As the most of the original edges remained, the generator derives not perfect but better anomaly images from Fig.\ref{fig:7}(1) and (3). It indicates that a suitable portion of anomaly edges is the key to generate high quality anomaly images.

\begin{figure}
  \centering
  \setlength{\abovecaptionskip}{0.cm}
    \includegraphics[width=0.95\linewidth]{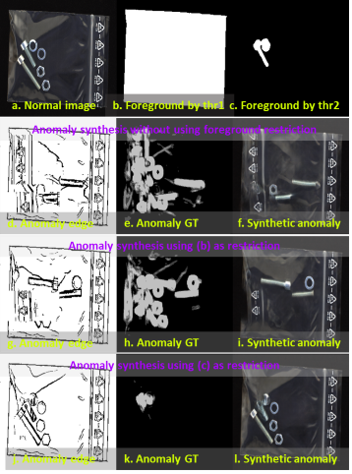}
  \caption{\textbf{Analysis of region selection.} Different foreground extractions may cause the inconsistency between original normal image and the synthetic anomaly image.}
  \label{fig:8}
  \vspace{-5mm}
\end{figure}

\textbf{Region selection.} Along with synthetic anomaly images, the ground truth for training anomaly localization module is simultaneously generated. Whether the ground truth conforms to logic is another issue that needs to be considered. Fig.\ref{fig:8} demonstrates examples of this problem. Given a normal image Fig.\ref{fig:8}(a), two possible generated anomaly images are shown in Fig.\ref{fig:8}(f)and (i) respectively. Due to largely replacement of original edges, the generated anomaly images of Fig.\ref{fig:6}(f) and (i) are almost completely unrelated with the input normal image Fig.\ref{fig:8}(a). Considering the underline logic of normal data, the ground truth of Fig.\ref{fig:8}(f) should indicate the missing parts base on Fig.\ref{fig:8}(f) rather than the Fig.\ref{fig:8}(a).
The anomaly localization module will be confused to learn from these kinds of training pairs.
An extreme example is the replacement of entire edge map with another one. 

\begin{figure*}
  \centering
  \setlength{\abovecaptionskip}{0.cm}
    \includegraphics[width=0.95\linewidth]{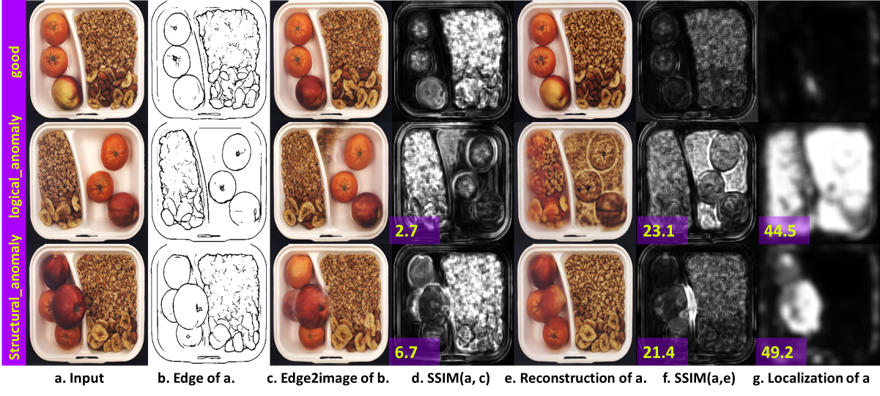}
  \caption{\textbf{Comparison of anomaly localization.} We compare our localization map (g) with the difference maps (d)(f) that are produced by calculating SSIM between generated images (c)(e) and input (a). (c) is generated by the edge-to-image generator. (e) is generated by our reconstruction sub-network. The pixel AUC-sPro of the class(breakfastBox) are illustrated in lower left corners.}
  \label{fig:9}
  \vspace{-5mm}
\end{figure*}

Different with above mentioned situations, the synthetic anomaly image shown in Fig.\ref{fig:8}(l) having the consistency with the normal image and can provide a reasonable ground truth Fig.\ref{fig:8}(k) for the anomaly localization training. The success of Fig.\ref{fig:8}(l) is due to its effective foreground restriction, as shown in Fig.\ref{fig:8}(c), extracted with a larger threshold(thr2$>$thr1). Due to edges caused by background reflection, such as plastic packaging, it is difficult to precisely extract the foreground from JND map with a simple and costless method.
The foreground extraction is not necessary to be perfect but have to remain parts of the key components to maintain consistency. Motivated by these observations, we construct anomaly edges based on region selection strategies that avoids generation collapse and maintains consistency.

\textbf{Anomaly localization.} Fig.\ref{fig:9} illustrates the advance performance of our localization sub-network comparing directly using SSIM metric to spot anomaly regions. Fig.\ref{fig:9}c and Fig.\ref{fig:9}d indicate that the edge-to-image generator converts edge maps (b) to color images (c) with little sense of anomalies. It is barely impossible to use it indicate anomaly regions, especially the logical anomaly. On the contrary, our reconstruction images (e) reveal more vivid corruption in the anomaly regions.\ref{fig:9}e.

\section{Loss functions}
\label{sec:lossfunction}
The total loss $L$ consists of reconstruction $L_{rec}$ and segmentation $L_{seg}$ losses. For both JND map and normal image reconstruction, we not only use MSE loss to supervise the pixel-to-pixel recovering but also the structural similarity (SSIM) \cite{ssim} loss to yield plausible local consistency.
\begin{equation}
L=L_{rec}+L_{seg}
\end{equation}
\begin{equation}
L_{rec}=L_{img}+L_{jnd}+L_{edge}
\end{equation}
\begin{equation}
L_{img}=L_2(I,I_{rec})+L_ssim(I,I_{rec})
\end{equation}
\begin{equation}
L_{jnd}=L_2(J,J_{rec})+L_ssim(J,J_{rec})
\end{equation}
where I and J are input normal image and corresponding JND map, $I_{rec}$ and $J_{rec}$ are output reconstructions.

Following PiDiNet \cite{pidinet}, we adopt the annotator-robust loss function proposed in \cite{edgeloss} for the reconstructed edge maps. For the ith pixel in the jth edge map with value $p_i^j$, the loss is calculated as:
\begin{equation}
L_{edge}=\left\{
\begin{aligned}
&\alpha\cdot(log(1-p^j)),if\ y=0 \\
&0,if\ 0<y<\eta \\
&\beta\cdot(logp^j),otherwise
\end{aligned}
\right.
\end{equation}
where y is the ground truth edge probability generated by PiDiNet \cite{pidinet}, $\eta$ is the binary threshold, $\beta=1.1$ is the percentage of negative pixel samples and $\alpha=1.0$ is the percentage of positive pixel samples.

To handle the unbalance of normal and anomaly, we use focal loss \cite{focalloss} to supervise the predicted anomaly localization.
\begin{equation}
L_{seg}=L_{focalloss}(M,M_{seg})
\end{equation}
\section{Anomaly synthesis}
\label{sec:anomalysynthesis}
\begin{figure*}
  \centering
  \setlength{\abovecaptionskip}{0.cm}
    \includegraphics[width=0.95\linewidth]{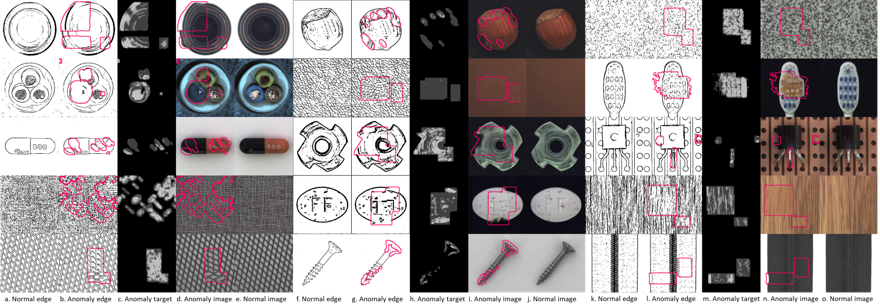}
  \caption{\textbf{Visualization of synthetic anomaly for MVTecAD \cite{mvtec}.}}
  \label{fig:10}
  \vspace{-1mm}
\end{figure*}

\begin{figure*}
  \centering
  \setlength{\abovecaptionskip}{0.cm}
    \includegraphics[width=0.95\linewidth]{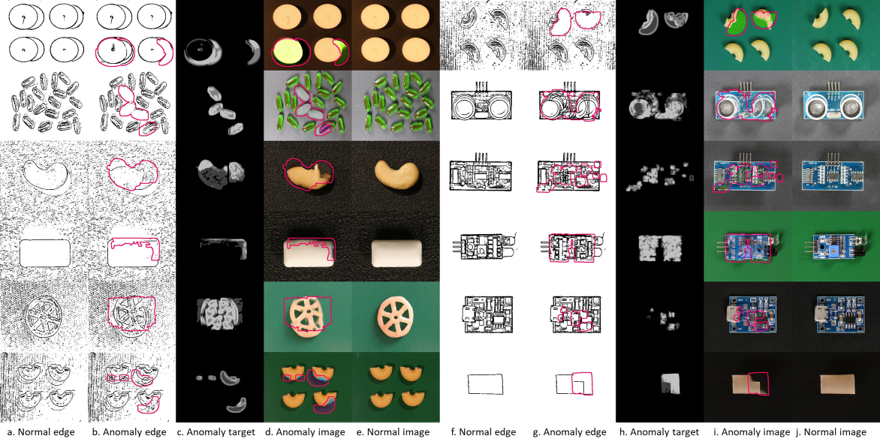}
  \caption{\textbf{Visualization of synthetic anomaly for VisA \cite{visa}.}}
  \label{fig:11}
  \vspace{-1mm}
\end{figure*}

\begin{figure*}
  \centering
  \setlength{\abovecaptionskip}{0.cm}
    \includegraphics[width=0.95\linewidth]{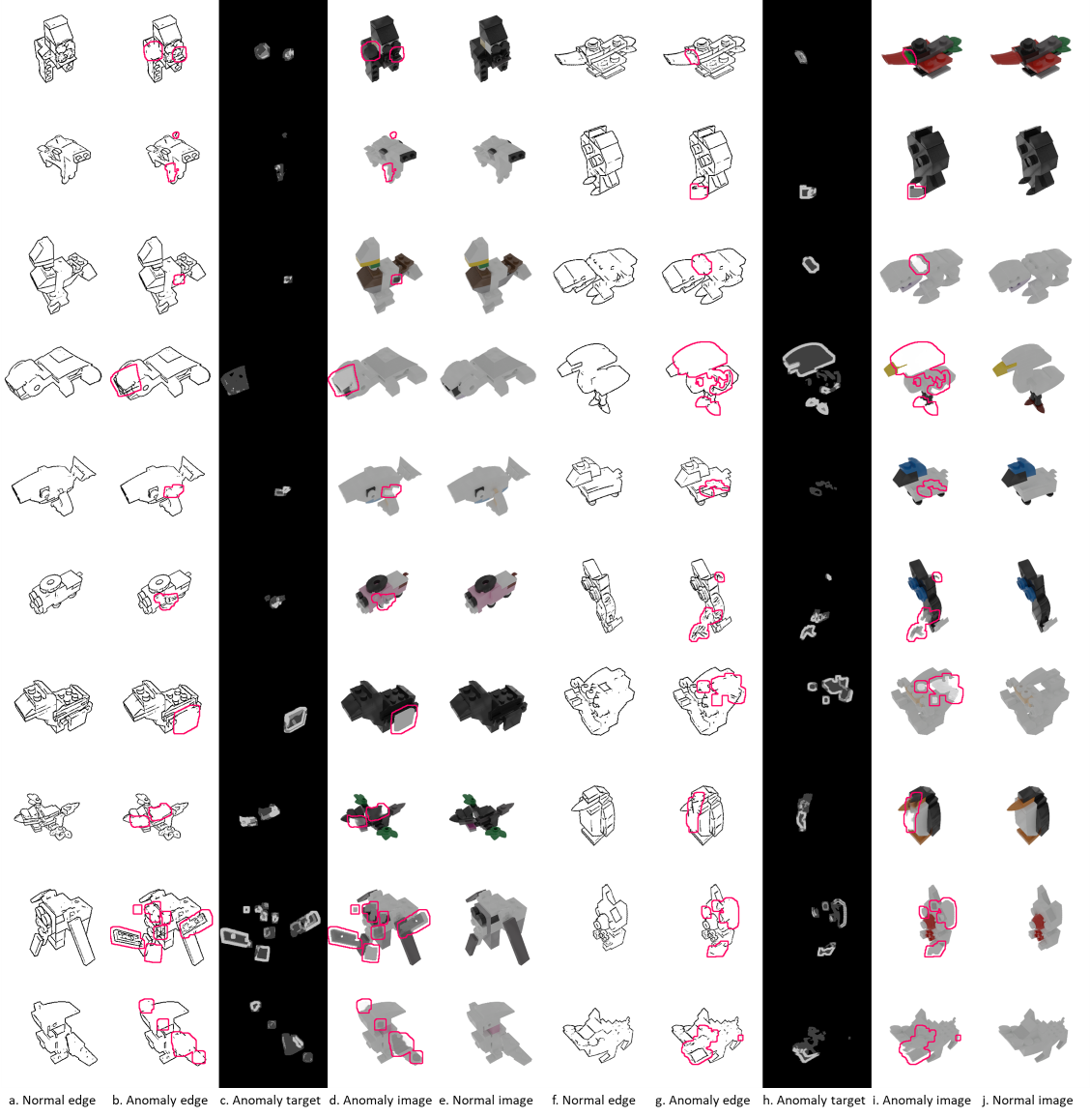}
  \caption{\textbf{Visualization of synthetic anomaly for MADsim \cite{pad}.}}
  \label{fig:12}
  \vspace{-1mm}
\end{figure*}

Fig.\ref{fig:10}-Fig.\ref{fig:12} visualize the synthetic anomalies of MVTecAD \cite{mvtec}, VisA \cite{visa} and MADsim \cite{pad} datasets respectively. We randomly apply region selection and edge modification strategies to generate anomaly edge maps for all datasets. We also randomly apply colorjitter to VisA \cite{visa} and TPS warping to MADsim \cite{pad} datasets.

\begin{figure*}
  \centering
  \setlength{\abovecaptionskip}{0.cm}
    \includegraphics[width=0.95\linewidth]{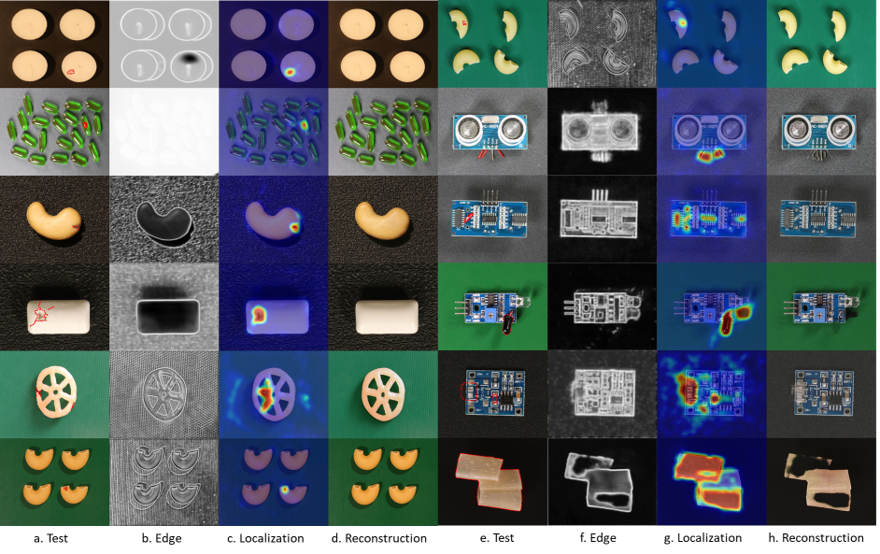}
  \caption{\textbf{Qualitative illustration of our anomaly detection results on VisA \cite{visa}.}}
  \label{fig:13}
  \vspace{-1mm}
\end{figure*}

\begin{figure*}
  \centering
  \setlength{\abovecaptionskip}{0.cm}
    \includegraphics[width=0.95\linewidth]{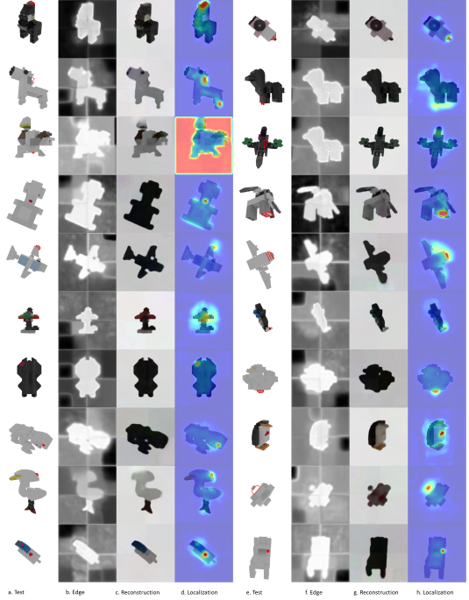}
  \caption{\textbf{Qualitative illustration of our anomaly detection results on MADsim \cite{pad}.}}
  \label{fig:14}
  \vspace{-1mm}
\end{figure*}

\section{Qualitative evaluation}
\label{sec:qualitativeevaluation}
Fig.\ref{fig:13}-Fig.\ref{fig:14} visualize the synthetic anomalies of VisA \cite{visa} and MADsim \cite{pad} datasets respectively.


\end{document}